\documentclass[final]{cvpr}
\usepackage{times}
\usepackage{epsfig}
\usepackage{graphicx}
\usepackage{amsmath}
\usepackage{amssymb}
\usepackage{enumitem}
\usepackage{pifont}
\usepackage{hhline}
\usepackage{url}

\usepackage{booktabs}
\usepackage{nicefrac}
\usepackage{microtype}
\usepackage{graphicx}
\usepackage{csquotes}
\usepackage{subfigure}
\usepackage{wrapfig}
\usepackage{multirow}
\usepackage{floatrow}
\usepackage{diagbox}
\usepackage{verbatim}
\usepackage{amsmath}
\usepackage{bbm}
\usepackage{bm}
\usepackage{comment}
\usepackage{amsthm}
\usepackage{amssymb}
\usepackage{amsmath}
\usepackage{nicefrac}
\usepackage{comment}
\usepackage{epigraph}
\usepackage{bbm}

\usepackage[misc]{ifsym}
\newtheorem{theorem}{Theorem}

\newtheorem{remark}{Remark}
\newtheorem{proposition}[theorem]{Proposition}

\numberwithin{theorem}{section}

\newcommand\blfootnote[1]{%
  \begingroup
  \renewcommand\thefootnote{}\footnote{#1}%
  \addtocounter{footnote}{-1}%
  \endgroup
}

\def\cvprPaperID{2486} 

\begin{document}

\title{Forecasting Irreversible Disease via Progression Learning}

\author{Botong~Wu$^{1,2*}$,\quad Sijie~Ren$^{7*}$,\quad Jing~Li$^{1,6}$,
\quad Xinwei~Sun$^{4 (\text{\Letter})}$,\quad  Shiming Li$^{5}$, \quad  Yizhou~Wang $^{1,3}$
\\
\textsuperscript{1} Dept. of  Computer Science, Peking University
\quad
\textsuperscript{2} Adv. Inst. of Info. Tech, Peking University \quad
\\
\textsuperscript{3}Center on Frontiers of Computing Studies, Peking University
\\
\textsuperscript{4}Microsoft Research, Asia
\quad
\textsuperscript{5}Beijing Tongren Hospital, Capital Medical University\\
\textsuperscript{6} Deepwise AI Lab \quad \textsuperscript{7}Beijing Stars Universal Technology Co., Ltd\\
{\tt\small
\{botongwu, lijingg, yizhou.wang\}@pku.edu.cn, rensijie@ooyby.com, xinsun@microsoft.com} 
}

\maketitle

\begin{abstract}
\blfootnote{* denotes equal contribution.}
Forecasting Parapapillary atrophy (PPA), i.e., a symptom related to most irreversible eye diseases, provides an alarm for implementing an intervention to slow down the disease progression at early stage. A key question for this forecast is: how to fully utilize the historical data (e.g., retinal image) up to the current stage for future disease prediction? In this paper, we provide an answer with a novel framework, namely \textbf{D}isease \textbf{F}orecast via \textbf{P}rogression \textbf{L}earning (\textbf{DFPL}), which exploits the irreversibility prior (i.e., cannot be reversed once diagnosed). Specifically, based on this prior, we decompose two factors that contribute to the prediction of the future disease: i) the current disease label given the data (retinal image, clinical attributes) at present and ii) the future disease label given the progression of the retinal images that from the current to the future. To model these two factors, we introduce the current and progression predictors in DFPL, respectively. In order to account for the degree of progression of the disease, we propose a temporal generative model to accurately generate the future image and compare it with the current one to get a residual image. The generative model is implemented by a recurrent neural network, in order to exploit the dependency of the historical data. To verify our approach, we apply it to a PPA in-house dataset and it yields a significant improvement (\textit{e.g.}, \textbf{4.48\%} of accuracy; \textbf{3.45\%} of AUC) over others. Besides, our generative model can accurately localize the disease-related regions.

\end{abstract}

\section{Introduction}


The World Health Organization (WHO) estimates that 19 million children below the age of 15 were visually impaired \cite{who-2004,who-ref-2019} (1\% of the total population in this age group). Most of the eye diseases, such as myopia in children \cite{li2020automatic}, glaucoma \cite{zhang2012automatic}, retinal detachment, and dense cataract \cite{holden2016global}, are highly related to Parapapillary atrophy (PPA), which as a biomarker of above eye diseases, refers to outer retinal atrophy adjacent to the optic disc \cite{teng2010beta, PPA-cur-2012, PPA-cur-2020}. Due to the irreversibility of these eye diseases, forecasted PPA can be provided as an alarm to implement an intervention (\textit{e.g.}, outdoor activities, or drug treatment) to prevent the rapid progression of eye diseases at the early stage. Due to the lack of future data when forecasting the future label, this forecasting task is equivalent to the following answer: \textit{how to fully utilize the longitudinal/sequential data up to the current stage for future disease prediction, under the lack of future data?}

A series of works have recently been proposed to answer this question, such as \cite{AD-follow-up-baseline-2015, follow-up-ipmi-2019, follow-up-AD-2018}. Most of these works utilized the provided current data for generating the future medical data (\textit{i.e.}, retinal images, clinical attributes), followed by an auxiliary classifier for disease prediction. However, these methods did not take the \emph{irreversibility} medical prior into account.

This irreversibility prior overlooked in the above literature refers to that, the disease cannot reverse to healthy once diagnosed. That is, if diagnosed as diseased at present, the probability of disease at the future stage would be 100\%. Inspired by such a \emph{prior} in PPA\cite{kim2018longitudinal}, we decompose (according to the law of total probability) the disease label at future stage into two factors: \textbf{i)} the disease label at current stage given the medical data at present; and \textbf{ii)} the disease label at future stage given the progression from the current to the future stage. This factorization, in contrast to previous works that only leverage current data for the generation, claims an additional role of current data in determining the disease at present (\textit{a.k.a the i)}). To effectively learn these two factors, we propose a novel framework, namely \textbf{D}isease \textbf{F}orecast via \textbf{P}rogression \textbf{L}earning (\textbf{DFPL}) which introduces two prediction modules: $f_{\mathrm{cur}}$ and $f_{\mathrm{prog}}$, respectively. To further account for the degree of progression, we propose a temporal generative framework based on Generative Adversarial Networks (GAN), in which we incorporate the generator with the recurrent neural network that takes prior sequential data as input to predict the feature map in the next stage. By comparing this generated feature map with the one at the current stage, one can get the residual feature map, as a measurement of the degree of progression.

\begin{figure*}
\vspace{-0.2cm}
    \includegraphics[width=0.9\linewidth]{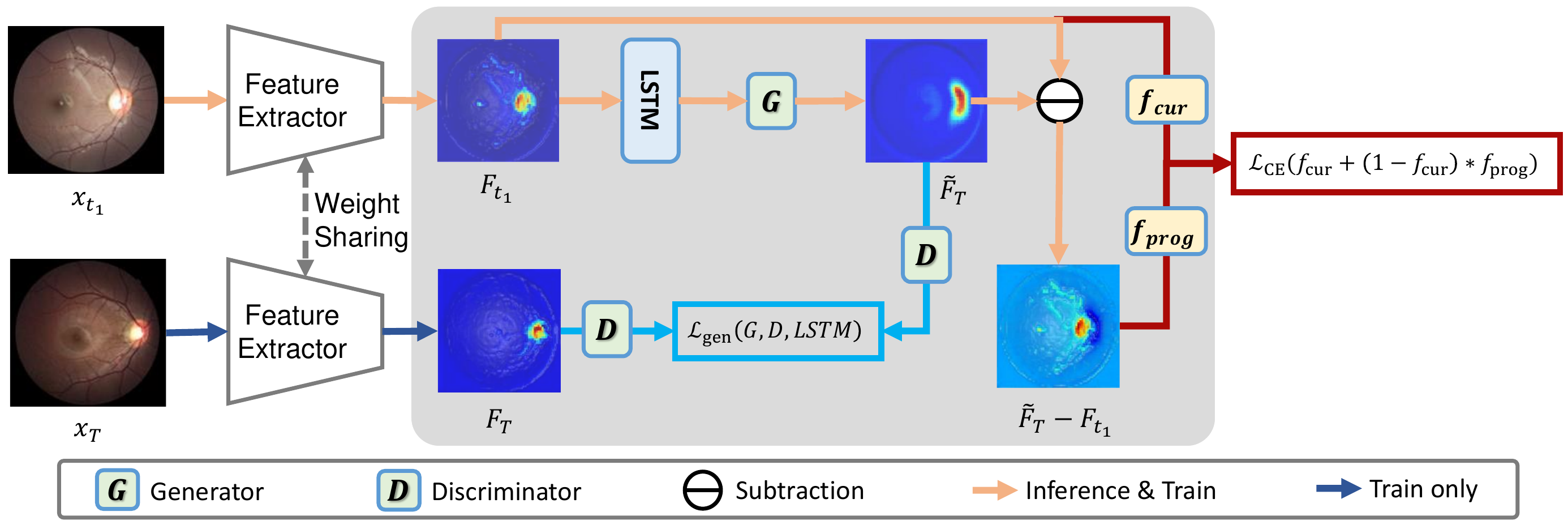}
    \caption{Illustration of our learning framework DFPL. We first pre-train a feature extractor and the extracted feature maps denoted as $\bm{F}_t$ (together with clinical attributes) are taken as inputs. The modules contained in the gray area are trained in an end-to-end scheme. Specifically the $\bm{F}_{T}$ generated by $\mathbb{G}(\mathrm{LSTM}(\bm{F}_{t_1},T-t_1), \bm{a}_{t_1})$ is trained to compete with discriminator $\mathbb{D}$ by adversarial loss. The final prediction is the combination of the current predictor $f_{\mathrm{cur}}$ and the residual predictor $f_{\mathrm{prog}}$ with residual feature maps $\tilde{\bm{F}}_{T} - \bm{F}_{t_1}$ as input.}
    \label{fig:frame}
    \vspace{-0.3cm}
\end{figure*}

To validate the utility of our approach, we apply it to a in-house data which belongs to a longitude PPA protocol for clinical diagnosis for primary-school-aged children. The results show a large improvement over others in terms of prediction accuracy (ACC) and Area Under the ROC Curve (AUC): \textit{e.g.}, \textbf{4.48\%} of accuracy; \textbf{3.45\%} of AUC. Besides, the visualization result shows that our DFPL equipped with temporal generative learning can localize the disease-related regions such as \textit{optic disc}. An ablation study is further conducted to verify the contribution of each module of our framework. The main contributions can be summarized as follows:

\begin{itemize}
    \item We are \textit{the first} to point out the two-fold effects from the longitudinal data up to the current stage to the forecast for irreversible diseases: the disease status at present and the one based on progression from the current stage. We propose a novel framework to learn such two effects.
    \item We propose a temporal generative framework equipped with a recurrent neural network, to learn the dynamics of disease progression.
    \item Our method can achieve better prediction results than others on an in-house PPA data of primary-school-aged children; besides, the detected disease-related regions can be concentrated on the optic disc.
\end{itemize}

\section{Related Work}

Forecasting disease with longitudinal data refers to predicting the disease label at a future stage, given the sequential data up to the current stage. As a simple and effective approach, the deterministic-type method  \cite{AD-MLP-BGRU-2018-ISBI,AD-RNN-long-2019-CMIG,OCT-treat-Dense-RNN-long-2020-JBHI} adopted a two-step strategy: first, they extracted semantic features using the convolutional neural network; then they fed these features into a recurrent neural network to predict the future outcome. Alternatively, due to the ability to capture the temporal relation among the sequential data during generation, a series of generative-based methods \cite{follow-up-AD-2018,follow-up-ipmi-2019} have recently been proposed. As a typical example, the \cite{follow-up-AD-2018} proposed to generate future data (\textit{e.g.} $T$) with the data at the current stage (\textit{e.g.} $t < T$), via generative adversarial networks \cite{goodfellow2014generative}; such a generated data, as a reflection of the progression from the current data to the future, was then fed into a classifier to predict the disease label. The \cite{follow-up-ipmi-2019} proposed to learn the smooth Riemannian manifold of the whole trajectory, from the low-dimensional latent space via the deep generative model. Compared to \cite{follow-up-AD-2018}, the \cite{follow-up-ipmi-2019} additionally leveraged the information from the past (\textit{e.g.} $\{\tilde{t} < t\}$). However, these methods did not exploit the irreversibility prior \cite{follow-up-AD-2018,follow-up-ipmi-2019} and the dependency among sequential data \cite{follow-up-AD-2018} during modeling.

\noindent \textbf{Our Specification.} Our method is better-motivated in that we exploit the fact that the disease cannot be reversed at any time in the future once diagnosed, to propose the two-fold effects for disease forecast: the current disease status and the progression. We formulate this proposition as a theoretical guideline of our learning framework, specifically the current and the progression predictors. To further account for the degree of progression, we propose a temporal GAN equipped with the recurrent neural network to generate the future feature maps; besides, we employ the high-order dynamics (\textit{e.g.}, first-order $\bm{x}_{t_2} - \bm{x}_{t_1}; \bm{x}_{t_3} - \bm{x}_{t_2}$ and second-order $(\bm{x}_{t_3} - \bm{x}_{t_2}) - (\bm{x}_{t_2} - \bm{x}_{t_1})$) as input for prediction.

\section{Methodology}

\noindent \textbf{Problem Setting \& Notation.} Our goal is to predict the disease label $y_T$ at future stage $T$, given (a subset of) retinal fundus images $\bm{x}_{\leq t}$ and clinical attributes $\bm{a}_{\leq t}$ (\textit{e.g., height, time for outdoor activities, myopia situation of parents, etc.}) at some time $t$ with $t < T$. The $y_t \in \{\pm 1\}$ for any $t > 0$, with $+1,-1$ respectively denoting the disease and healthy status, without loss of generality. Our data for training this classifier contain $N$ subjects: $\{\bm{s}^i\}_{i \in [N]}$, where $[N] := \{1,...,N\}$ and $\bm{s}^i = \left( \bm{z}^i_{\leq t}, y^i_T \right)$ where $\bm{z}^i_t = \left(\bm{x}^i_{t} ,\bm{a}^i_{t} \right)$. Note that due to labeling cost, we do not require the labels $y_t$ before $T$ (\textit{i.e.}, $t < T$), except at initial time point $t=1$ such that all samples are healthy, i.e., $y^i_1 = -1$ for all $i \in [N]$. We denote the $K$-order setting as the data of $K$ time points are provided for training and testing, \textit{i.e.}, $(\bm{x}_{t_{1:K}},\bm{a}_{t_{1:K}})$.

\noindent \textbf{Outline.} We first introduce our roadmap in section~\ref{sec:roadmap}, guided by our finding that the future disease is affected by the current stage and the progression, as formulated in Prop.~\ref{prop:prog}. We then introduce our learning framework in section~\ref{sec:framework}, with each module detailedly explained. Finally, we generalize our method to high-order cases (multiple points of images and attributes are observed) in section~\ref{sec:high-order}.

\subsection{Roadmap with Theoretical Guideline}
\label{sec:roadmap}

We consider the disease forecast, \textit{i.e.}, $p(y_{T}=1|\bm{z}_{t_{1:K}})$ with $t_{1}<...<t_{K}<T$ denoting a sequence of $K$ time points. For simplicity, in the following we consider 1-order case with $K=1$ (with high-order case $K>1$ introduced in section~\ref{sec:high-order}). Due to the inability of reverse the disease status without medical treatment \cite{lu2011quantification}, this future prediction should satisfy the following principles:
\begin{itemize}[topsep=0pt,itemsep=-1ex]
  \item  \textit{Irreversibility}: Once diagnosed as PPA, one would not transfer to healthy in the future, if no medical intervention is implemented.
  \item  \textit{Deterioration}: The probability of PPA is monotonic with respect to the time $t$.
\end{itemize}
Based on the \textit{Irreversibility} principle, it can be induced that the disease status in the next stage is affected by \textit{(i)} the situation in the current stage and \textit{(ii)} the progression speed as time grows, which is formulated as the following proposition:
\begin{proposition}
\label{prop:prog}
Under the \textit{irreversibility} principle, we have the following factorization for progression prediction:
\begin{align}
    \label{eq:prog}
     p(y_{T} =1|\bm{z}_{t_1}) = & \underbrace{p(y_{t_1} =1 | \bm{z}_{t_1})}_{\mathrm{Current}} + \nonumber \\
     p(y_{t_1} =0|\bm{z}_{t_1})  & \underbrace{p(y_{T} =1| y_{t_1} = 0, \bm{z}_{t_1})}_{\mathrm{Progression}}.
\end{align}
\end{proposition}
\begin{remark}
The Prop~\ref{prop:prog} shows that $p(y_{T}=1|\bm{z}_{t_1}) \geq p(y_{t_1}=1|\bm{z}_{t_1})$, agreeing with the \textit{Deterioration} principle.
\end{remark}
As a guideline, we can correspondingly design two modules to respectively model the current disease prediction and the dynamic progression. Besides, the term ``progression" can be obtained by
\begin{align}
\label{eq:prog-term}
& p(y_{T} =1| y_{t_1} = 0, \bm{z}_{t_1}) = \\
& \int_{\bm{x}_{T}} p(\bm{x}_{T}|y_{t_1}=0,\bm{z}_{t_1})* p(y_{T}=1|y_{t_1}=0,\bm{x}_{T},\bm{z}_{t_1}) d\bm{x}_{T}. \nonumber
\end{align}
For $p(y_{T}=1|{y}_{t_{1}}=0,\bm{x}_{T},\bm{z}_{t_{1}})$ that describes the extent of progression from the healthy status, we propose to approximate it using progression information which contains \textit{i.e.}, $\bm{x}_{T} - \bm{x}_{t_1}$. We summarize the above conclusions as a roadmap for our learning framework.

\noindent \textbf{RoadMap.} We first pre-train a feature extractor to extract feature maps $\bm{F}_{t_1}$ from retinal fundus images $\bm{x}_{t_1}$. And the future feature maps $\tilde{\bm{F}}_T$ are estimated by a trainable temporal generative model with extracted $\bm{F}_{t_1}$. Then, we learn two prediction modules: $f_{\mathrm{cur}}$ and $f_{\mathrm{prog}}$, respectively with the feature maps $\bm{F}_{t_1}$ and the residual feature maps $\tilde{\bm{F}}_{T} - \bm{F}_{t_{1}}$ as inputs. The residual feature maps are calculated to measure the degree of progression, as which the current feature maps $\bm{F}_{t_{1}}$ are subtracted from the estimated feature maps in future stage $\tilde{\bm{F}}_{T}$. We will introduce our learning framework in details in the subsequent section. 

\subsection{Disease Forecast via Progression Learning}
\label{sec:framework}

We introduce our learning framework, namely \textbf{D}isease \textbf{F}orecast via \textbf{P}rogression \textbf{L}earning (DFPL), with high-level spirit stated in the roadmap in the above section. In more detail, as illustrated in Fig.~\ref{fig:frame}, we first pre-train a feature extractor $\mathrm{Enc}$ to extract feature maps from image at each time point. With extracted feature maps at different time steps, \textit{i.e.}, $\bm{F}_{t_1},...,\bm{F}_{t_K}$ (here we set $K=1$ for simplicity), we train a convolutional Long Short-Term Memory (LSTM) \cite{hochreiter1997long} followed by a generator $\mathbb{G}$ to generate the next stage feature maps, in an adversarial way via Generative Adversarial Networks (GAN) \cite{goodfellow2014generative}. The extracted feature maps at current stage (\textit{i.e.} $\bm{F}_{t_1}$) and the residual feature maps with estimated feature maps at the future stage (\textit{i.e.}, $\tilde{\bm{F}}_{T} - \bm{F}_{t_1}$), are respectively taken as inputs for classification modules $f_{\mathrm{cur}}$ and $f_{\mathrm{prog}}$. The final prediction is given by $f_{\mathrm{cur}} + (1 - f_{\mathrm{cur}})  f_{\mathrm{prog}}$, which is optimized via cross-entropy loss in empirical risk minimization. In the following, we will explain all these modules in details: the pre-trained feature extractor $\mathrm{Enc}$; generative model which is composed of generator $\mathbb{G}$, discriminator $\mathbb{D}$ and the recurrent neural network (here we adopt LSTM \cite{hochreiter1997long}); current predictor $f_{\mathrm{cur}}$ and progression predictor $f_{\mathrm{prog}}$.

\noindent \textbf{Pre-trained Feature Extractor ($\mathrm{Enc}$).} Instead of training directly on images, we implement a pre-training strategy to obtain feature maps denoted as $\bm{F}$ as the input of classifiers (together with attributes $\bm{a}$), which has been found to be effective in the literature \cite{erhan2010does}. Specifically, we train a classifier on \textbf{(i)} $\{\bm{x}^i_t,y^i_t\}_{t \in \{1,T\},i\in [N]}$ (recall that $y_{t=1}^i = -1$ for all $i$) to extract features representative of current disease status; and on \textbf{(ii)} $\{\bm{x}^i_t,y^i_T\}_{t < T, i \in [N]}$ to extract features that related to the progression. The bottom layers of neural networks after pre-training are  (\textit{e.g.}, the first two blocks for ResNet18 in experiment) denoted as feature extractor $\mathrm{Enc}$. In the following, we take extracted feature maps as input of modules LSTM, $\mathbb{G}$, $\mathbb{D}$, $f_{\mathrm{cur}}$ and $f_{\mathrm{prog}}$ (the gray area in Fig.~\ref{fig:frame}).

\noindent \textbf{Generative Model.} The goal is to generate the feature maps at future stage (\textit{i.e.}, $\tilde{\bm{F}}_T$). By comparing it with the $\bm{F}_{t_1}$ at the current stage, the $\tilde{\bm{F}}_T - \bm{F}_{t_1}$ measures the degree of progression and is thus fed into $f_{\mathrm{prog}}$ to predict the $p(y_T=1|{y}_{t_1}=0,\bm{z}_{t_1})$ in Eq.~\eqref{eq:prog}. For an accurate generation, we adopt the adversarial training strategy, specifically the Wasserstein GAN (WGAN) \cite{wgan} with weight clipping, to train the generator $\mathbb{G}$ and a discriminator $\mathbb{D}$ in a competing way. To further capture the dependency of the historical feature maps, we additionally train a LSTM of which the output is then fed into the generator $\mathbb{G}$, as shown in Fig.~\ref{fig:frame}. The generative loss function for the 1-order generation (with the higher-order generation introduced later) is $\mathcal{L}_{\mathrm{gen}}(\mathbb{G},\mathbb{D},\mathrm{LSTM})$. The generative loss is computed by the real future feature maps $\bm{F}^i_{T}$ and the generated future feature maps $\tilde{\bm{F}}^i_{T} = \mathbb{G}(\mathrm{LSTM}(\bm{F}^i_{t_1},T - t_1),\bm{a}^i_{t_1})$.

\noindent \textbf{Current Predictor.} The $f_{\mathrm{cur}}$, as the predictor of current disease status given $\bm{F}_t := \mathrm{Enc}(\bm{x}_t)$\footnote{The input does not contain $\bm{a}$, since it is unobserved at $T$ for the 2nd term in Eq.~\eqref{eq:cur-erm}}, is trained via the empirical risk minimization (ERM) of labeled training data (initial time point and future time point) and generated future data:
\begin{align}
\label{eq:cur-erm}
    \mathcal{L}_{\mathrm{ERM}}(f_{\mathrm{cur}}) & = \sum_{i\in [N]} \left( \sum_{t \in \{1,T\}} \log{\frac{1}{p_{f_{\mathrm{cur}}}(y^i_t|\bm{F}^i_{t},\bm{a}^i_{t})}} + \right.\nonumber \\
    &  \left.  \sum_{t_1 <T}\log{\frac{1}{p_{f_{\mathrm{cur}}}(y^i_T|\tilde{\bm{F}}^i_T(\bm{F}^i_{t_1}, \bm{a}^i_{t_1}))}} \right),
\end{align}
where $\tilde{\bm{F}}^i_T(\bm{F}^i_{t_1},\bm{a}^i_{t_1}) := \mathbb{G}(\mathrm{LSTM}(\bm{F}^i_{t_1},T-t_1),\bm{a}^i_{t_1})$. Therefore, the $\mathcal{L}_{\mathrm{ERM}}$ also trains the generator $\mathbb{G}$ and the $\mathrm{LSTM}$, which is omitted here for simplicity. Besides, we additionally regularize $p({y}_{T} = 1|\bm{F}_{T}) \geq p({y}_{t_1} = 1|\bm{F}_{t_1})$ for any $T > t_1$ according to the \textit{Deterioration} principles, formulated as soft-margin regularization:
\begin{align}
\label{eq:cur-regularize}
    \mathcal{J}_{\mathrm{cur}}(f_{\mathrm{cur}})
    = \! \sum_{i\in[N], t_1<T} \! \max\left(0, \mathrm{diff}_i(\bm{F}^i_{T}, \bm{F}^i_{t_1}) + \theta \right),
\end{align}
where $\mathrm{diff}_i(\bm{F}^i_{T}, \bm{F}^i_{t_1}) := p_{f_{\mathrm{cur}}}(y^i_{t_1} = 1|\bm{F}^i_{t_1}) -p_{f_{\mathrm{cur}}}(y^i_{T} = 1|\bm{F}^i_{T})$ and $\theta > 0$ denotes the margin hyper-parameter. The overall loss function to train $f_{\mathrm{cur}}$ is:
\begin{equation}
\label{eq:cur-loss}
    \mathcal{L}_{\mathrm{cur}}(f_{\mathrm{cur}}) = \mathcal{L}_{\mathrm{ERM}}(f_{\mathrm{cur}}) + \alpha \mathcal{J}_{\mathrm{cur}}(f_{\mathrm{cur}}),
\end{equation}
with $\alpha > 0$ denoting the hyper-parameter that balances the effects of prediction and the \textit{Deterioration} principle.

\noindent \textbf{Progression Predictor.} As aforementioned in sec.~\ref{sec:roadmap}, the ``progression" term can be approximated by $p({y}_T=1|{y}_{t_1}=0,\bm{z}_{t_1}) \approx$ $\int p(\bm{F}_T|\bm{F}_{t_1},\bm{z}_{t_1}) p({y}_T=1|\bm{F}_T-\bm{F}_{t_1}, \bm{a}_{t_1})d\bm{F}_T$. The loss for $f_{\mathrm{prog}}$ taking the residual feature maps $\tilde{\bm{F}}_{T} - \bm{F}_{t_1}$ as input and also $f_{\mathrm{cur}}$, according to factorization of ``current" and ``progression" term in Prop.~\ref{prop:prog}, is reformulated as:
\begin{align}
    & \mathcal{L}_{\mathrm{CE}}(f_{\mathrm{prog}},f_{\mathrm{cur}}) = \! \! \sum_{i\in[N], t_1 < T} \! \! \log{\frac{1}{p_{f_{\mathrm{cur}},f_{\mathrm{prog}}}(y_T^i|\tilde{\bm{F}}^i_{T},\bm{F}^i_{t_1},\bm{a}^i_{t_1})}}, \label{eq:res} \\
    \vspace{-0.1cm}
    & p_{f_{\mathrm{cur}},f_{\mathrm{prog}}}(y_T^i=1|\tilde{\bm{F}}^i_{T},\bm{F}^i_{t_1},\bm{a}^i_{t_1}) =  p_{f_{\mathrm{cur}}}(y_{t_1}^i=1|\bm{F}^i_{t_1},\bm{a}^i_{t_1}) + \nonumber \\
    & p_{f_{\mathrm{cur}}}(y_{t_1}^i\!=\!0|\bm{F}^i_{t_1},\bm{a}^i_{t_1}) p_{f_{\mathrm{prog}}}(y_T^i \!=\!1|\tilde{\bm{F}}_{T} - \bm{F}_{t_1},\bm{a}^i_{t_1}). \label{eq:cur-res}
\end{align}
Note that the $\mathcal{L}_{\mathrm{CE}}$ also depends on the generator $\mathbb{G}$ and the $\mathrm{LSTM}$ since that the $\tilde{\bm{F}}_{T} := \mathbb{G}(\mathrm{LSTM}(\bm{F}_{t_1},T-t_1),\bm{a}_{t_1})$.

\noindent \textbf{Training $\&$ Inference.} Combining separate losses for the modules mentioned above (specifically Eq.~\eqref{eq:cur-loss}, $\mathcal{L}_{\mathrm{gen}}(\mathbb{G},\mathbb{D},\mathrm{LSTM})$ and Eq.~\eqref{eq:res}), the overall loss function is defined as:
\begin{align}
    \label{eq:loss}
   & \mathcal{L}(f_{\mathrm{cur}},f_{\mathrm{prog}},\mathbb{G},\mathbb{D},\mathrm{LSTM}) := \mathcal{L}_{\mathrm{gen}}(\mathbb{G},\mathbb{D},\mathrm{LSTM})  \nonumber \\
    & \quad +  \lambda_1 * \mathcal{L}_{\mathrm{cur}}(f_{\mathrm{cur}})  + \lambda_2 * \mathcal{L}_{\mathrm{CE}}(f_{\mathrm{prog}},f_{\mathrm{cur}}).
\end{align}
During inference, given $(\bm{x}_{t_1},\bm{a}_{t_1})$, we first obtain $\bm{F}_{t_1}$ via $\mathrm{Enc}(\bm{x}_{t_1})$. Then we generate the $\tilde{\bm{F}}_{T}$ via $\mathbb{G}(\mathrm{LSTM}(\bm{F}_{t_1},T-t_1),\bm{a}_{t_1})$. Then we feed $(\tilde{\bm{F}}_{T},\bm{F}_{t_1},\bm{a}_{t_1})$ into $p_{f_{\mathrm{cur}},f_{\mathrm{prog}}}$ in Eq.~\eqref{eq:cur-res} for prediction.

\subsection{Extension to High-Order Prediction}
\label{sec:high-order}

We extend our loss in Eq.~\eqref{eq:loss} to leverage high-order information (including the information from the past, \textit{i.e.}, $\bm{z}_{t_{1:K-1}}$ and the current, \textit{i.e.}, $\bm{z}_{t_{K}}$) into the generation of feature maps at future stage and hence the future disease, \textit{i.e.}, $p(y_{T}=1|\bm{z}_{t_{1:K}})$ with $K>1$. The Prop.~\ref{prop:prog} for this case is presented similarly, with factorization of the current and the progression (please refer to supplementary for details). Therefore, the whole framework can be inherited and the extensions of $K$-order for current predictor, generative model and progression predictor are summarized as follows. 

\noindent \textbf{Current Predictor.} We consider the $p({y}_{t_K}|\bm{z}_{t_{1:K}})$ for any $t_1<..<t_K < T$. To leverage the information before $t_K$, \textit{i.e.}, $\bm{z}_{t_{1:K-1}}$, we additionally train a classifier from $\bm{z}_t$ to ${y}_T$ (the label only given at $T$), namely $f_{\mathrm{fut}}$ (with ``fut" standing for the word ``future"):
\begin{align}
\label{eq:future}
\mathcal{L}_{\mathrm{fut}}(f_{\mathrm{fut}}) = \sum_{i \in [N], t_1<T} \frac{1}{\log{p_{f_{\mathrm{fut}}}(y^i_T|\bm{F}^i_{t_1},\bm{a}^i_{t_1})}}.
\end{align}
Based on the current predictor $f_{\mathrm{cur}}$ and future predictor $f_{\mathrm{fut}}$, the $p({y}_{t_K}|\bm{F}_{t_{1:K}},\bm{a}_{t_{1:K}})$ is then modeled as:
\begin{align}
\label{eq:cur-high}
    & \quad \quad p_{f_{\mathrm{cur}},f_{\mathrm{fut}}}({y}_{t_K}|\bm{F}_{t_{1:K}},\bm{a}_{t_{1:K}}) =  \\
    & \frac{1}{K} \left( p_{f_{\mathrm{cur}}}({y}_{t_K}|\bm{F}_{t_K}, \bm{a}_{t_K}) + \sum_{j=1}^{K-1} p_{f_{\mathrm{fut}}}({y}_{t_K}|\bm{F}_{t_j},\bm{a}_{t_j})  \right). \nonumber
\end{align}

\noindent \textbf{Generative Model.} To leverage the high-order information into the generation of the future maps in $T$, from the past $K$-length sequence (\textit{i.e.} $\bm{F}_{t_1},\bm{a}_{t_1},...,\bm{F}_{t_K},\bm{a}_{t_K}$ for any $t_1<t_2<...<t_K < T$), we iteratively feed the feature maps and related attributes into the LSTM up to the $t_K$, followed by the generator $\mathbb{G}$ that is trained by adversarial loss to compete the discriminator $\mathbb{D}$. The $\mathcal{L}_{\mathrm{gen}}(\mathbb{G},\mathbb{D},\mathrm{LSTM})$ is computed by the real feature maps and generated feature maps from $t_2$ to $T$. Equipped with the LSTM's ability of long-term memory, this high-order generation can capture the time-dependency.

\noindent \textbf{Progression Predictor.} For progression learning, the high-order residual information can be approximated by differentiation of the ones of lower-order (\textit{e.g.}, the second-order residual at time $t_2$ can be approximated by difference of two first-order residuals $\bm{F}_{t_3} - \bm{F}_{t_2}$ and $\bm{F}_{t_2} - \bm{F}_{t_1}$ as $(\bm{F}_{t_3} - \bm{F}_{t_2}) - (\bm{F}_{t_2} - \bm{F}_{t_1})$). Generally speaking, the set of $\{j\}_{j \in [K]}$-order residual information denoted as $\mathrm{prog}_K(\tilde{\bm{F}}_{T},\{\bm{F}_{t_j}\}_{j \in [K]})$ is composed of \textbf{(i)} the first-order information $\{\tilde{\bm{F}}_{T}-\bm{F}_{t_{K}},\{\bm{F}_{t_{K-i}}-\bm{F}_{t_{K-i-1}}\}_{i=0}^{K-2} \}$; and \textbf{(ii)} the ones related to the $j$-th order for $j \geq 2$, represented by $\{(\tilde{\bm{F}}_{T}- \bm{F}_{t_{K+2-j}}) - (\bm{F}_{t_{K}}- \bm{F}_{t_{K+1-j}}), \{ (\bm{F}_{t_{K-i}}- \bm{F}_{t_{K-i+1-j}}) - (\bm{F}_{t_{K-i-1}}- \bm{F}_{t_{K-i-j}}) \}_{i=0}^{K-2} \}$. The loss is the same with Eq.~\eqref{eq:res} except that the input of $f_{\mathrm{prog}}$ turns to $\mathrm{prog}_K(\tilde{\bm{F}}_{T},\{\bm{F}_{t_j}\}_{j \in [K]})$ ($K > 1$) and $\bm{a}_{t_{1:K}}$, and the $p_{f_{\mathrm{cur}}}(\bm{F}_{t_1})$ is replaced with Eq.~\eqref{eq:cur-high} that additionally leverage the information . In summary, the $p_{f_{\mathrm{cur}},f_{\mathrm{prog}}}$ in Eq.~\eqref{eq:cur-res} (with $\bm{\zeta}: = \{\bm{F},\bm{a}\}$) is replaced by:
\begin{align}
    \label{eq:cur-res-high}
    & p_{f_{\mathrm{cur}},f_{\mathrm{prog}},f_{\mathrm{fut}}}({y}_T^i=1|\bm{\zeta}^i_{t_{1:K}}) = \nonumber \\
    & \quad p_{f_{\mathrm{cur}},f_{\mathrm{fut}}}({y}_{t_K}=1|\bm{\zeta}^i_{t_{1:K}}) + p_{f_{\mathrm{cur}},f_{\mathrm{fut}}}({y}_{t_K}=0|\bm{\zeta}^i_{t_{1:K}}) \nonumber \\
    & \quad p_{f_{\mathrm{prog}}}({y}_T^i=1|\mathrm{prog}_K(\tilde{\bm{F}}^i_{T},\{\bm{F}^i_{t_j}\}_{j \in [K]}),\bm{a}^i_{t_{1:K}}).
\end{align}

\noindent \textbf{Training $\&$ Inference.} The overall loss function on high-order setting is same as 1-order one Eq.\eqref{eq:loss} except that the future loss Eq.\eqref{eq:future} need to be considered and the prediction for $\mathcal{L}_{\mathrm{CE}}$ is computed by Eq.~\eqref{eq:cur-res-high}. The overall loss function on high-order setting is defined as:
\begin{align}
    \label{eq:high-loss}
   & \mathcal{L}(f_{\mathrm{cur}},f_{\mathrm{prog}},f_{\mathrm{fut}},\mathbb{G},\mathbb{D},\mathrm{LSTM}) := \mathcal{L}_{\mathrm{gen}}(\mathbb{G},\mathbb{D},\mathrm{LSTM}) \nonumber \\
    & +  \lambda_1 * \mathcal{L}_{\mathrm{cur}}(f_{\mathrm{cur}}) + \lambda_2 * \mathcal{L}_{\mathrm{CE}}(f_{\mathrm{prog}},f_{\mathrm{cur}},f_{\mathrm{fut}}) \nonumber \\
    & +  \lambda_3 * \mathcal{L}_{\mathrm{fut}}(f_{\mathrm{fut}}).
\end{align}
During inference, the process is the same to 1-order setting except that feeding a sequential data $(\bm{x}_{t_{1:K}},\bm{a}_{t_{1:K}})$ into $\mathrm{Enc}$, $\mathbb{G}$ and $\mathrm{LSTM}$ to compute the feature maps $\bm{F}_{t_{1:K}}$ the high-order residual information set $\mathrm{prog}_K(\tilde{\bm{F}}^i_{T},\{\bm{F}^i_{t_j}\}_{j \in [K]})$. Then we feed above feature maps and related attributes into $f_{\mathrm{cur}},f_{\mathrm{prog}}$ and $f_{\mathrm{fut}}$ in Eq.~\eqref{eq:cur-res-high} for prediction. 


\begin{table*}[]
  \centering
  \caption{The ACC, AUC (mean $\pm$ std) comparisons between our method and the baselines on the 1-order setting. $\delta t = T - t_i$ with the $\delta t = 1$ implying that the input of the test sample is from 5th graders since $T$ represents $T = 6$. Average over ten runs.}
  \vspace{-0.2cm}
  \renewcommand\tabcolsep{1.0pt}
  \small
  \resizebox{\textwidth}{15.5mm}{
    \begin{tabular}{|c|c|c|c|c|c|c|c|c|c|c|}
    \hline
    Methods & \multicolumn{2}{c|}{RN18} & \multicolumn{2}{c|}{MM-F} & \multicolumn{2}{c|}{ARL} & \multicolumn{2}{c|}{TCSL} & \multicolumn{2}{c|}{Ours} \\
    \hline
    Num of Param & \multicolumn{2}{c|}{138.68M} & \multicolumn{2}{c|}{137.19M} & \multicolumn{2}{c|}{138.69M} & \multicolumn{2}{c|}{157.51M} & \multicolumn{2}{c|}{141.56M} \\
    \hline
    Metric & \multicolumn{1}{c|}{ACC} & \multicolumn{1}{c|}{AUC} & \multicolumn{1}{c|}{ACC} & \multicolumn{1}{c|}{AUC} & \multicolumn{1}{c|}{ACC} & \multicolumn{1}{c|}{AUC} & ACC   & AUC   & \multicolumn{1}{c|}{ACC} & \multicolumn{1}{c|}{AUC} \\
    \hline
    $\delta t$=5 & 60.53$\pm$2.55 & 63.56$\pm$2.51 & 63.46$\pm$1.77 & 64.38$\pm$1.62 & 63.14$\pm$1.90 & 65.58$\pm$3.24 & 58.56$\pm$0.56 & 56.53$\pm$1.01 & \textbf{66.67}$\pm$1.94 & \textbf{72.37}$\pm$0.82 \\
    \hline
    $\delta t$=4 & 65.25$\pm$2.01 & 70.41$\pm$1.98 & 66.38$\pm$2.50 & 71.88$\pm$1.54 & 67.88$\pm$1.92 & 73.88$\pm$2.33 & 62.98$\pm$0.96 & 66.89$\pm$0.64 & \textbf{69.80}$\pm$1.94 & \textbf{76.88}$\pm$0.42 \\
    \hline
    $\delta t$=3 & 62.80$\pm$2.77 & 66.80$\pm$1.64 & 63.07$\pm$1.65 & 67.52$\pm$2.36 & 67.09$\pm$1.02 & 71.32$\pm$2.52 & 66.12$\pm$1.69 & 69.20$\pm$0.92 & \textbf{69.98}$\pm$1.69 & \textbf{78.65}$\pm$1.02 \\
    \hline
    $\delta t$=2 & 69.92$\pm$1.92 & 75.78$\pm$2.56 & 69.91$\pm$2.34 & 79.48$\pm$1.88 & 70.80$\pm$2.68 & 80.13$\pm$1.19 & {74.52}$\pm$1.54 & 79.65$\pm$0.45 & \textbf{77.16}$\pm$1.28 & \textbf{83.52}$\pm$1.12 \\
    \hline
    $\delta t$=1 & 73.05$\pm$3.36 & 82.74$\pm$1.51 & 75.50$\pm$2.86 & 86.70$\pm$1.04 & 75.30$\pm$3.02 & 86.68$\pm$1.57 & 77.53$\pm$1.94 & 84.88$\pm$0.27 & \textbf{79.37}$\pm$2.09 & \textbf{87.16}$\pm$1.12 \\
    \hline
        Average & 66.31$\pm$1.24 & 71.86$\pm$1.20 & 67.6.7$\pm$0.78 & 73.99$\pm$0.76 & 68.84$\pm$1.47 & 75.52$\pm$1.52 & 67.88$\pm$1.18 & 71.43$\pm$0.20 & \textbf{72.60}$\pm$1.53 & \textbf{79.72}$\pm$0.50 \\
    \hline
    \end{tabular}}%
  \label{tab:order_1}%
  \vspace{-0.2cm}
\end{table*}%

\begin{table*}[]
  \centering
  \caption{The ACC, AUC (mean $\pm$ std) comparisons between our method and the baselines on the 2-order setting. $\delta t = T - t_i$ with the $\delta t = 1$ implying that the input of test samples are from 4th graders and 5th graders. Average over ten runs.}
  \vspace{-0.2cm}
  \renewcommand\tabcolsep{1.0pt}
  \small
  \resizebox{\textwidth}{15.5mm}{
    \begin{tabular}{|c|c|c|c|c|c|c|c|c|c|c|}
    \hline
    Methods & \multicolumn{2}{c|}{RN18} & \multicolumn{2}{c|}{MM-F} & \multicolumn{2}{c|}{ARL} & \multicolumn{2}{c|}{RGL} & \multicolumn{2}{c|}{Ours} \\
    \hline
    Num of Param & \multicolumn{2}{c|}{154.90M} & \multicolumn{2}{c|}{155.46M} & \multicolumn{2}{c|}{154.91M} & \multicolumn{2}{c|}{150.28M} & \multicolumn{2}{c|}{152.72M} \\
    \hline
    Metric & \multicolumn{1}{c|}{ACC} & \multicolumn{1}{c|}{AUC} & \multicolumn{1}{c|}{ACC} & \multicolumn{1}{c|}{AUC} & \multicolumn{1}{c|}{ACC} & \multicolumn{1}{c|}{AUC} & ACC   & AUC   & \multicolumn{1}{c|}{ACC} & \multicolumn{1}{c|}{AUC} \\
    \hline
    $\delta t$=4 & 62.54$\pm$1.81 & 68.09$\pm$1.67 & 68.84$\pm$1.80 & 74.21$\pm$1.35 & 67.96$\pm$2.24 & 74.23$\pm$1.47 & 69.24$\pm$1.25 & 76.34$\pm$1.25  & \textbf{70.17}$\pm$1.11 & \textbf{76.42}$\pm$1.20 \\
    \hline
    $\delta t$=3 & 64.81$\pm$1.62 & 71.47$\pm$1.70 & 71.22$\pm$1.92 & 78.38$\pm$0.87 & 71.13$\pm$1.37 & 77.49$\pm$2.01 & 68.39$\pm$1.68   & 74.38$\pm$1.26  & \textbf{72.75}$\pm$1.69 & \textbf{80.16}$\pm$0.52 \\
    \hline
    $\delta t$=2 & 67.07$\pm$3.53 & 74.07$\pm$1.39 & 73.43$\pm$1.86 & 81.14$\pm$0.67 & 70.99$\pm$0.62 & 80.15$\pm$2.51 & 75.26$\pm$2.14  & 81.98$\pm$0.64  & \textbf{77.17}$\pm$1.69 & \textbf{85.28}$\pm$0.51 \\
    \hline
    $\delta t$=1 & 73.43$\pm$2.32 & 81.84$\pm$0.93 & 76.57$\pm$1.19 & 85.80$\pm$0.93 & 75.42$\pm$1.49 & 85.34$\pm$1.49 & 78.33$\pm$1.95  & 87.71$\pm$1.17 & \textbf{78.64}$\pm$1.77 & \textbf{90.04}$\pm$0.27 \\
    \hline
    Average & 66.96$\pm$1.47 & 73.87$\pm$0.99 & 72.51$\pm$1.04 & 79.88$\pm$0.57 & 71.37$\pm$1.10 & 79.30$\pm$1.74 & 72.81$\pm$0.98 & 80.10$\pm$0.57 & \textbf{74.68}$\pm$1.25 & \textbf{82.98}$\pm$0.52 \\
    \hline
    \end{tabular}}%
  \label{tab:order_2}%
  \vspace{-0.4cm}
\end{table*}%

\begin{table*}[]
  \centering
  \caption{The ACC, AUC (mean $\pm$ std) comparisons between our method and the baselines on the 3-order setting. $\delta t = T - t_i$ with the $\delta t = 1$ implying that the input of the test samples is from 3rd graders, 4th graders, and 5th graders. Average over ten runs.}
  \vspace{-0.2cm}
  \renewcommand\tabcolsep{1.0pt}
  \small
  \resizebox{\textwidth}{13mm}{
    \begin{tabular}{|c|c|c|c|c|c|c|c|c|c|c|}
    \hline
    Methods & \multicolumn{2}{c|}{RN18} & \multicolumn{2}{c|}{MM-F} & \multicolumn{2}{c|}{ARL} & \multicolumn{2}{c|}{RGL} & \multicolumn{2}{c|}{Ours} \\
    \hline
    Num of Param & \multicolumn{2}{c|}{154.90M} & \multicolumn{2}{c|}{155.46M} & \multicolumn{2}{c|}{154.91M} & \multicolumn{2}{c|}{150.28M} & \multicolumn{2}{c|}{152.72M} \\
    \hline
    Metric & \multicolumn{1}{c|}{ACC} & \multicolumn{1}{c|}{AUC} & \multicolumn{1}{c|}{ACC} & \multicolumn{1}{c|}{AUC} & \multicolumn{1}{c|}{ACC} & \multicolumn{1}{c|}{AUC} & ACC   & AUC   & \multicolumn{1}{c|}{ACC} & \multicolumn{1}{c|}{AUC} \\
    \hline
    $\delta t$=3 & 63.77$\pm$1.74 & 68.72$\pm$1.55 & 67.36$\pm$1.17 & 73.29$\pm$1.96 & 68.50$\pm$2.90 & 74.98$\pm$2.78 & 66.04$\pm$2.51  & 71.69$\pm$1.09 & \textbf{74.03}$\pm$0.56 & \textbf{81.47}$\pm$0.52 \\
    \hline
    $\delta t$=2 & 68.67$\pm$2.34 & 74.41$\pm$1.16 & 73.54$\pm$1.29 & 80.26$\pm$1.20 & 70.07$\pm$2.67 & 80.07$\pm$2.19 & 73.57$\pm$1.73 & 79.81$\pm$1.48 & \textbf{77.53}$\pm$1.15 & \textbf{85.81}$\pm$0.51 \\
    \hline
    $\delta t$=1 & 75.69$\pm$1.64 & 80.33$\pm$1.32 & 75.30$\pm$2.22 & 84.51$\pm$1.06 & 74.55$\pm$3.09 & 85.48$\pm$1.08 & 77.31$\pm$1.31 & 85.84$\pm$1.15 & \textbf{79.56}$\pm$1.46 & \textbf{88.92}$\pm$0.31 \\
    \hline
    Average & 69.38$\pm$1.06 & 74.49$\pm$1.04 & 72.07$\pm$0.81 & 79.35$\pm$1.20 & 71.15$\pm$2.57 & 80.01$\pm$1.72 & 72.30$\pm$1.02  & 79.12$\pm$0.96 & \textbf{77.04}$\pm$1.01 & \textbf{85.40}$\pm$0.17 \\
    \hline
    \end{tabular}}%
  \label{tab:order_3}%
  \vspace{-0.2cm}
\end{table*}%

\begin{table*}[]
\caption{Ablation study on 1-order setting, to validate the effectiveness of each module. The Eq.\eqref{eq:prog} means that we train the model with loss Eq.\eqref{eq:loss} and predict by $f_{\mathrm{cur}} + (1 - f_{\mathrm{cur}}) f_{\mathrm{prog}}$. ``MA" stands for Model Average with $f_{\mathrm{cur}}(\tilde{\bm{F}}_T)$, Eq.\eqref{eq:prog} and $f_{\mathrm{fut}}(\bm{F}_{t_i})$. $f_{\mathrm{cur}}$ denotes that we train the model with $\mathcal{L}_{\mathrm{gen}} + \lambda_1 \mathcal{L}_{\mathrm{cur}}$ and predict by $f_{\mathrm{cur}}(\tilde{\bm{F}}_T)$. $f_{\mathrm{prog}}$ denotes that we train the model with $\mathcal{L}_{\mathrm{gen}} + \lambda_2' (- \log p_{f_{\mathrm{prog}}}(y_T|\tilde{\bm{F}}_{T} - \bm{F}_{t_1},\bm{a}_{t_1}))$ and predict by $f_{\mathrm{prog}}(\tilde{\bm{F}}_T - \bm{F}_{t_1}, \bm{a}_{t_1})$. }
\renewcommand\tabcolsep{4.0pt}
\vspace{-0.2cm}
\small
\begin{tabular}{|c|c|c|c|c|c|c|c|c|c|c|c|c|c|c|}
\hline
\multirow{2}{*}{Predictor} & \multirow{2}{*}{LSTM} & \multirow{2}{*}{MA} & \multicolumn{2}{c|}{$\delta t$=5} & \multicolumn{2}{c|}{$\delta t$=4} & \multicolumn{2}{c|}{$\delta t$=3} & \multicolumn{2}{c|}{$\delta t$=2} & \multicolumn{2}{c|}{$\delta t$=1} & \multicolumn{2}{c|}{Average}    \\ \cline{4-15}
                      &                       &                     & ACC             & AUC             & ACC             & AUC             & ACC             & AUC             & ACC             & AUC             & ACC             & AUC             & ACC            & AUC            \\ \hline
$f_{\mathrm{cur}}$    & $\surd$               & $\times$            & 64.64           & 63.73           & 67.96           & 75.52           & 66.85           & 73.64           & 71.82           & 83.02           & 79.56           & 88.03           & 70.17          & 76.79          \\ \hline
$f_{\mathrm{prog}}$   & $\surd$               & $\times$            & 66.85           & 69.33           & 66.85           & 70.10           & 68.51           & 73.93           & 70.17           & 77.75           & 74.03           & 79.82           & 69.28          & 74.19          \\ \hline
Eq.\eqref{eq:prog}                  & $\surd$               & $\times$            & \textbf{69.06}  & 71.60           & 69.61           & 75.83           & 69.06           & 76.54           & 72.38           & 82.47           & 78.45           & 86.70           & 71.71          & 78.63          \\ \hline
Eq.\eqref{eq:prog}                  & $\times$              & $\surd$             & 68.51           & 69.87           & 62.98           & 71.90           & 65.19           & 73.23           & 70.17           & 78.83           & 76.80           & 85.29           & 68.73          & 75.82          \\ \hline
Eq.\eqref{eq:prog}                  & $\surd$               & $\surd$             & 68.51           & \textbf{72.14}  & \textbf{71.82}  & \textbf{77.36}  & \textbf{71.82}  & \textbf{78.49}  & \textbf{77.90}  & \textbf{83.56}  & \textbf{81.77}  & \textbf{87.38}  & \textbf{74.36} & \textbf{79.79} \\ \hline
\end{tabular}
\label{tab:ablation_compare}
\vspace{-0.3cm}
\end{table*}

\begin{table}[htbp]
  \centering
  \caption{Comparisons of different $K$-order (time steps) leveraged. All settings share the same set of sample indexes.}
  \vspace{-0.2cm}
  \renewcommand\tabcolsep{3.0pt}
  \small
    \begin{tabular}{|c|c|c|c|c|c|c|}
    \hline
    Task & \multicolumn{2}{c|}{1-order} & \multicolumn{2}{c|}{2-order} & \multicolumn{2}{c|}{3-order} \\
    \hline
    Metric & ACC   & AUC   & ACC   & AUC   & ACC   & AUC \\
    \hline
    $\delta t$=3 & 65.75  & 74.69  & 68.51  & 76.79  & \textbf{71.82} & \textbf{78.95} \\
    \hline
    $\delta t$=2 & 69.06  & 79.88  & 74.59  & 82.45  & \textbf{75.69} & \textbf{83.37} \\
    \hline
    $\delta t$=1 & 75.14  & 84.26  & 76.24  & \textbf{88.64} & \textbf{77.35} & 87.49  \\
    \hline
    Average & 69.98  & 79.61  & 73.11  & 82.63  & \textbf{74.95} & \textbf{83.27} \\
    \hline
    \end{tabular}
  \label{tab:ablation-3-task}
  \vspace{-0.1cm}
\end{table}%

\begin{figure*}
    \vspace{-0.2cm}
    \includegraphics[width=0.88\linewidth]{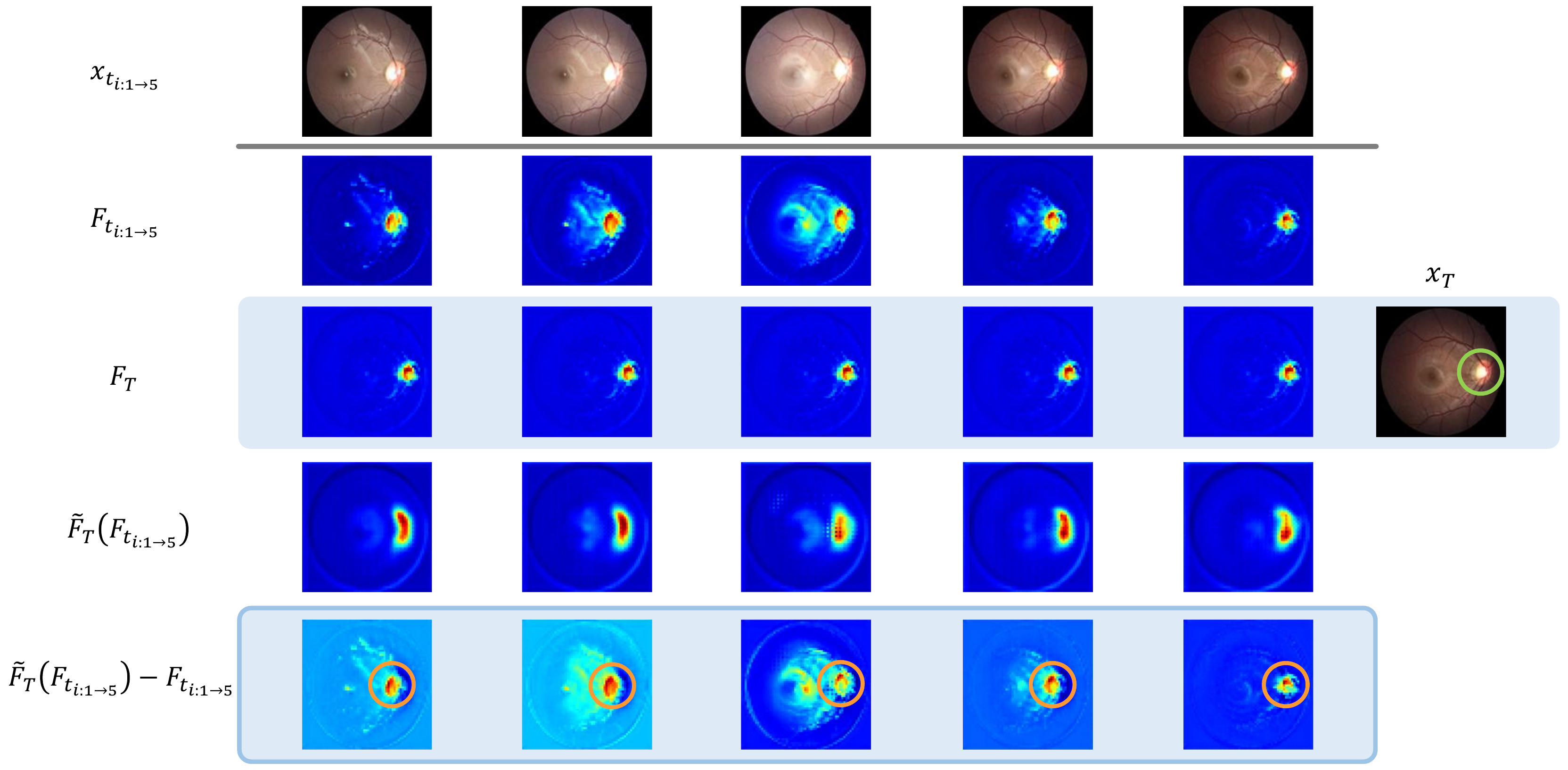}
    \caption{Visualization of learned feature maps on 1-order task. Feature maps from top to bottom are: $\bm{x}_{t_{i:1\to 5}}$ that represents retinal funds images, $\bm{F}_{t_{i:1\to 5}}$ denoting the feature maps extracted from images, the $F_T$, the $\tilde{\bm{F}}_T(\bm{F}_{t_{i:1 \to 5}})$ denoting the estimated feature maps via our generative model and the $\tilde{\bm{F}}_T(\bm{F}_{t_{i:1 \to 5}}) - \bm{F}_{t_{i:1 \to 5}}$ denoting the residual information.}
    \label{fig:frame-1}
    \vspace{-0.3cm}
\end{figure*}

\begin{figure*}[]
	\vspace{-0.2cm}
	\includegraphics[width=0.88\linewidth]{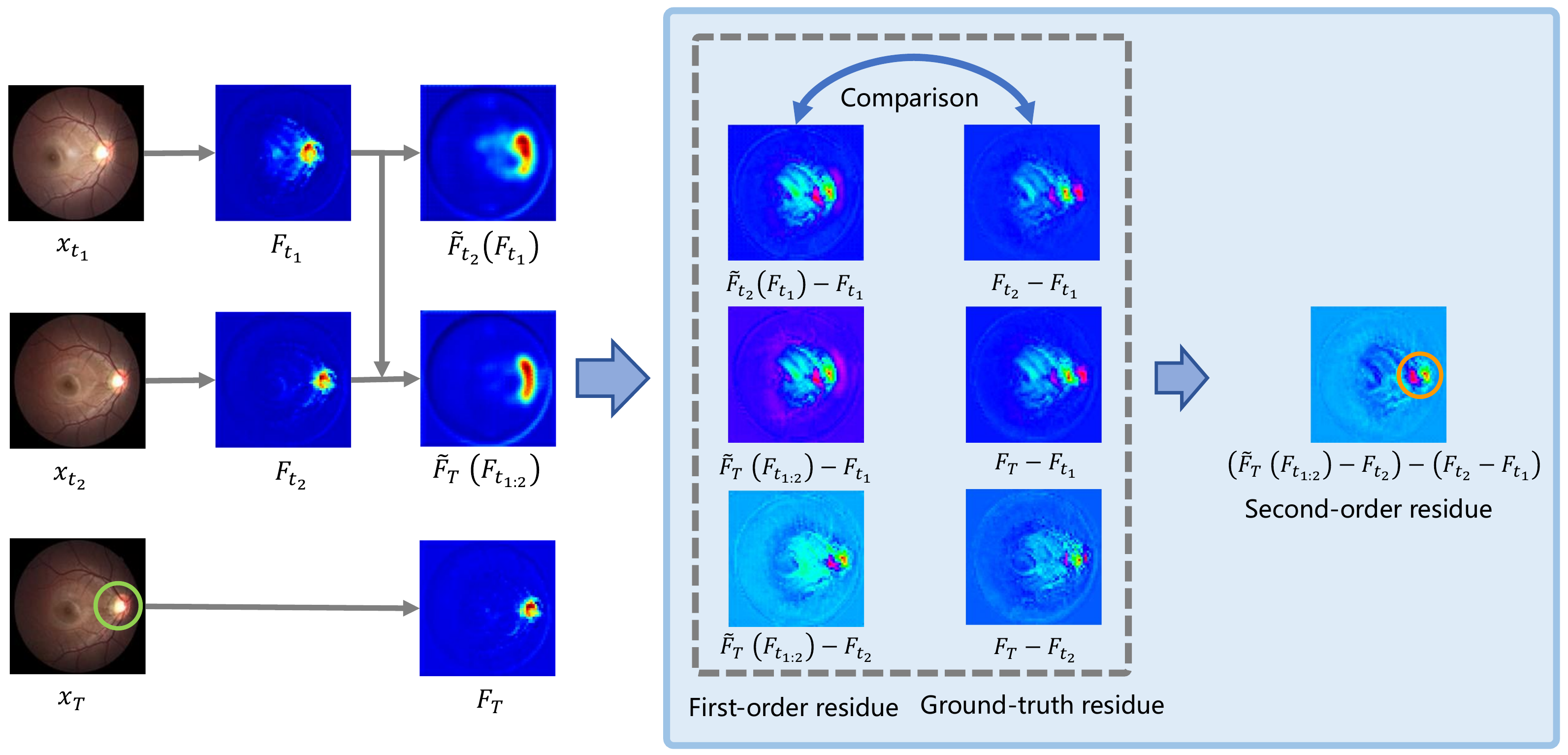}
	\caption{Visualization of estimated feature maps on 2-order task. Feature maps from left to right are:
		$\bm{x}_{t_{1:2}},\bm{x}_T$ denoting retinal funds images, $\bm{F}$ denoting the corresponding feature maps, $\tilde{\bm{F}}_{t_2}(\bm{F}_{t_1}), \tilde{\bm{F}}_{T}(\bm{F}_{t_{1:2}})$ denoting the estimated feature maps, estimated first-order residual feature maps, first-order residual feature maps and estimated second-order residual feature maps. We use red circle to mark the high-response area (the response extent is from high to low for the color from the red to the blue).
	}
	\label{fig:frame-2}
	\vspace{-0.3cm}
\end{figure*}

\section{Experimental Results}

In this section, we evaluate our method on an in-house longitudinal dataset, which studies the PPA progression for primary-school-aged (from grade-1 to grade-6) children.

\subsection{Dataset}

The data contains 905 participants in primary school, with each containing 3-6 data records (retinal fundus images $\bm{x}$ and clinical information $\bm{a}$, \textit{e.g.} height, time for outdoor activities, myopia situation of parents \footnote{For details please refer to supplementary information.}). In total, there are 5,046 data. Due to the costly labeling process, only the labels for the images from 1st graders and 6th graders are provided, with all participants at grade-1 being healthy. The data is randomly split into 60\% for training (543), 20\% for validation (181), and 20\% for testing (181) according to the index of participants. Our goal is to predict whether one would develop the disease at the final stage (\textit{i.e.} at grade-6), for any samples in the test data at the early stage.

\subsection{Baselines for Comparison}

a) \textbf{R}es\textbf{N}et-18 (RN18) \cite{resnet} which is trained to minimize cross entropy loss from $\bm{x}_{t}$ to ${y}_T$ for any $t < T$. For the network structure, we replace the $7 \times 7$ kernels in the first convolutional layer replaced by the two convolutional layers with kernel size $3 \times 3$. We empirically find that this replacement can achieve better prediction results. For simplicity, we name it as RN18, without otherwise specified.

b) \textbf{M}ulti-\textbf{M}odality-\textbf{F}using (MM-F) \cite{baltruvsaitis2018multimodal}, which proposed to fuse information of images $\bm{x}$ and clinical information $\bm{a}$ by concatenating features extracted from $\bm{x}$ via RN18 and those from $\bm{a}$ via a 3-layer (128$\to$256$\to$256) multilayer perceptron (MLP). It is also trained by minimizing cross entropy loss from $\bm{x}_{t}$ to ${y}_T$ for any $t < T$.

c) \textbf{T}emporal \textbf{C}orrelation \textbf{S}tructure \textbf{L}earning (TCSL) \cite{follow-up-AD-2018}. It implemented GAN to learn the joint distribution of $(\bm{x}_{t},\bm{x}_{t+1},{y}_{t+1})$ in order to capture the temporal relation between the adjacent points, followed by a classifier to predict the future label. Besides, it additionally trained a regression network to learn $\bm{x}_{t+1}$ from $\bm{x}_t$. For fair comparison, we adopt the same network structure
ofthe generator and the discriminator as ours. We adopt the RN18 for the follow-up classifier and the U-Net \cite{UNet-2D} for the regression network. More implementation details can refer to \cite{follow-up-AD-2018}. Since it generated only with adjacent point, we only compare it with our method on 1-order setting.

d) \textbf{A}ttention \textbf{R}esidual \textbf{L}earning (ARL) \cite{Res-2019}. It introduced an ARL-block which fuses input feature maps, residual feature maps, and attention feature maps to replace the traditional residual block in ResNet. The attention feature maps are computed by element-wise product of input feature maps and normalized residual feature maps. We replace the residual block with ARL-blocks for the MM-F method.

e) \textbf{R}iemannian \textbf{G}eometry \textbf{L}earning (RGL) \cite{follow-up-ipmi-2019}. It proposed a Riemannian manifold for the whole trajectory. As a high-order method, it implemented a deep generative model to map low-dimensional latent space to the high-dimensional observational data that lie in a geodesics of the manifold. We adopt the RN18 as the encoder and the same network structure of our generator as the decoder.

To compare with the high-order version of our method, we extend RN18, MM-F and ARL baselines to $K$-order version ($K>1$). Specifically, as for $K$-order method, we optimize the sum of cross entropy losses with the k-th loss taking $\bm{x}_{t_k}$ as input. The final prediction is $\frac{1}{K}\sum_{k=1}^K p(y_T=1|\bm{x}_{t_k})$.

\subsection{Implementation details}

We first pre-train a RN18 from $\bm{x}_{t}$ to ${y}_{t}$ for $t \in \{1,T\}$ and from $\bm{x}_{t}$ to ${y}_{T}$ for $t < T$, to obtain the feature extractor $\mathrm{Enc}$ as the first two convolutional layers followed by two residual blocks. The output of the feature extractor is 128 feature maps with size $64 \times 64$. Then we concatenate features from (i) down-sampled $32 \times 32$ feature maps via two Conv-BN-ReLU blocks \footnote{The ``Conv", ``BN" respectively stand for Convolution and Batch Normalization.}; and (ii) up-sampled $32 \times 32$ feature maps obtained from four TransposeConv-BN-ReLU blocks (with the channel size: $106\to2048\to1024\to1024\to512$) with a concatenated vector of clinical attributes $\bm{a} \in \mathbb{R}^6$ and a 100-dimensional Gaussian noise vector. The concatenated feature maps are then fed into a one-layer convolutional LSTM (with channel size 256) to generate the feature maps with size $32 \times 32$ at the next time point, followed by a generator $\mathbb{G}$ with a TransposeConv-BN-ReLU block and a Conv-BN-ReLU block (with the channel size: $256 \to 256 \to 128$) that outputs 128 feature maps with size $64 \times 64$ (same as the size of extracted feature maps). The discriminator $\mathbb{D}$ composes of five Conv-BN-LeakyReLU blocks (with the channel size $128\to256\to512\to1024\to1024\to1024\to1$). The negative slope of LeakyReLU is set to 0.2. As for $K$-order ($K > 1$) version, the input (and output) of LSTM module are changed to the corresponding sequential data with length of K.

We adopt center-cropping on the original image and resize them to $256 \times 256$. Then, we apply random rotation with $\leq 30$ degrees on each training image. All images are normalized with mean of 0.5 and std of 0.5. We respectively adopt RMSprop (with learning rate (lr) of 0.0001, weight decay (wd) of 0.0001) to train the generator and the discriminator and SGD (lr of 0.02, wd of 0.0001) to train the classification networks. We train the full model for 120 epochs and decay the lr by 0.2 every 60 epochs. The batch size is set to 20. The epoch number is optimized via the prediction accuracy on the validation set. The $\lambda_1$ and $\lambda_2$ in Eq.\eqref{eq:loss} are set to 0.1 and 1.0 for all order settings. The $\alpha$ in Eq.\eqref{eq:cur-loss} is set to 0.1. The $\lambda_3$ in Eq.\eqref{eq:high-loss} is set to 1.0.  During inference, we ensemble the models: i) Eq.~\eqref{eq:cur-res-high}, ii) $p_{f_{\mathrm{cur}}}$ in Eq.~\eqref{eq:cur-high} and iii) $p_{f_{\mathrm{fut}}}$ in Eq.~\eqref{eq:future}, \textit{i.e.}, $\frac{1}{K+2}\left( p_{f_{\mathrm{cur}},f_{\mathrm{prog}},f_{\mathrm{fut}}}({y}_{T}=1|{\bm{F}}_{t_{1:K}},\bm{a}_{t_{1:K}}) \right.$+ $\left. p_{f_{\mathrm{cur}}}({y}_{T}=1|\tilde{\bm{F}}_{T}, \bm{0}) + \sum_{j=1}^{K} p_{f_{\mathrm{fut}}}({y}_{T}=1|\bm{F}_{t_j}, \bm{a}_{t_j}) \right)$. $\bm{0} \in \mathbb{R}^6$ denotes the zero vector due to the attributes are not given at the future stage $T$. The average and standard deviation over 10 runs are reported.

\vspace{-0.2cm}
\subsection{Quantitative Results}

We consider three evaluation settings: 1-order in Tab.~\ref{tab:order_1}, 2-order in Tab.~\ref{tab:order_2} and 3-order in Tab.~\ref{tab:order_3}. The TCLS \cite{follow-up-AD-2018}, which only leveraged adjacent point for generation is only compared with others on 1-order setting; and the RGL \cite{follow-up-ipmi-2019} which generates the trajectory is compared with on 2-order and 3-order settings. As shown in Tab.~\ref{tab:order_1},\ref{tab:order_2} and~\ref{tab:order_3}, our method perform better and comparable than others in terms of prediction accuracy (ACC) and AUC metrics on all settings.

\vspace{-0.2cm}
\subsection{Ablation Study}
We conduct an ablation study to validate the effectiveness of each module. The results are summarized in Tab.~\ref{tab:ablation_compare}. As shown, the improvement of Eq.~\eqref{eq:prog} (the 3rd row) over the first two rows (validate the effectiveness of $f_{\mathrm{cur}},f_{\mathrm{prog}}$ in the disease forecast, as guided by Prop.~\ref{prop:prog}. Besides, the incorporation of LSTM into our model can bring additional improvement (of the 5th row over the 4th row), due to the ability of LSTM to exploit the dependency of sequential data. Finally, implementing the model ensemble can achieve further improvement, as shown by the result in the 5th row compared to the one in the 3rd row.

Moreover, to validate the advantage of leveraging higher-order information to generate future image (hence residual feature map), we keep the samples with data on all-time steps ($t=1:6$) provided for 1-order, 2-order, and 3-order settings. As shown in Tab.~\ref{tab:ablation-3-task}, the higher-order data we leverage, the better performance we can achieve (0.64\% AUC of 3-order over 2-order; and 3.02\% AUC of 2-order over 1-order).

\subsection{Visualization}
To verify that our method can learn interpretable features for disease forecast, we visualize the estimated feature maps by our method on 1-order and 2-order settings in Fig.~\ref{fig:frame-1} and Fig.~\ref{fig:frame-2}, respectively. In Fig.~\ref{fig:frame-1} for one diseased case, the feature maps from top to bottom are, real images from 1st graders to 5th graders ($t_{i: 1 \to 5}$), feature maps generated by feature extractor $\mathrm{Enc}$ on images from 1st graders to 5th graders (\textit{i.e.}, $\bm{F}_{t_{i: 1 \to 5}}$), the 5-time repeated feature maps generated by $\mathrm{Enc}$ on image from 6th graders (\textit{i.e.}, $\bm{F}_{T}$), the future feature maps (\textit{i.e.}, $\tilde{\bm{F}}_{T=6}(\bm{F}_{t_{i:1\to 5}})$ estimated by our temporal generative model which respectively taking $\bm{F}_{t_1}$,...,$\bm{F}_{t_5}$ as inputs,  residual feature maps $\tilde{\bm{F}}_{T} - \bm{F}_{t_{i:1\to 5}}$. In Fig.~\ref{fig:frame-2}, from left to right are: real images (\textit{i.e.}, $\bm{x}_{t_1}, \bm{x}_{t_2}, \bm{x}_T$ ($t_1 < t_2 < T$)), the corresponding feature maps via $\mathrm{Enc}$ (\textit{i.e.}, $\bm{F}_{t_1}, \bm{F}_{t_2}, \bm{F}_T$ ($t_1 < t_2 < T$)), the estimated feature maps by our generative model (\textit{i.e.}, $\tilde{\bm{F}}_{t_2}(\bm{F}_{t_1}), \tilde{\bm{F}}_{T}(\bm{F}_{t_{1:2}})$), the progression information in orange box which from left to right are: first-order estimated residual feature maps $\tilde{\bm{F}}_{t_2}(\bm{F}_{t_1}) - \bm{F}_{t_1}$, $\tilde{\bm{F}}_{T}(\bm{F}_{t_{1:2}}) - \bm{F}_{t_1}$, $\tilde{\bm{F}}_{T}(\bm{F}_{t_{1:2}}) - \bm{F}_{t_2}$ first-order residual feature maps $\bm{F}_{t_2} - \bm{F}_{t_1}$, $\bm{F}_{T} - \bm{F}_{t_1}$, $\bm{F}_{T} - \bm{F}_{t_2}$; second-order estimated residual feature maps: $(\tilde{\bm{F}}_{T}(\bm{F}_{t_{1:2}}) - \bm{F}_{t_{2}}) - (\bm{F}_{t_{2}} - \bm{F}_{t_{1}})$.

As shown in both Fig.~\ref{fig:frame-1} and~\ref{fig:frame-2}, the high response regions (marked by the orange circle) in learned residual feature maps (the last row marked by the blue rectangle in Fig.~\ref{fig:frame-1}, the last column in orange box in Fig.~\ref{fig:frame-2}) are concentrated in the optic disc region (marked by the green circle in the third row of Fig.~\ref{fig:frame-1}) which has been found to be highly correlated with PPA \cite{kim2012optic, song2018progressive}. Besides, it can be shown from the second row in Fig.~\ref{fig:frame-1} that the high-response regions in feature maps for 1st graders to 5th graders are more concentrated and similar to that of $\bm{F}_T$, which matches with our \textit{Deteriorate principle}. Another interesting phenomena, as shown in Fig.~\ref{fig:frame-2}, is that the high-response area in $\tilde{\bm{F}}_{T} - \bm{F}_{t_2}$ (also $\bm{F}_{T} - \bm{F}_{t_2}$) is smaller than that of $\tilde{\bm{F}}_{T} - \bm{F}_{t_1}$ (also $\bm{F}_{T} - \bm{F}_{t_1}$), which validates the interpretability of our residual images in describing the degree of progression information.


\vspace{-0.2cm}
\section{Conclusions \& Discussions}
\vspace{-0.3cm}
We present a framework to perform Disease Forecast via Progression Learning (DFPL) applied on an in-house sequential dataset for Parapapillary Atrophy forecast. To our knowledge, we are \textit{the first} to identify the two-fold effects (disease at present and the progression) for disease forecasting. The high-order residual information is employed to achieve a more accurate prediction result. Equipped with a recurrent neural network in our temporal generative model, the disease-related region is localized accurately. In the future, we will apply our method to other irreversible diseases, such as Alzheimer's Disease.

\section{Acknowledgements}
This work was supported in part by MOST-2018AAA0102004, NSFC-61625201 and NSFC-62061136001.

\newpage
{\small
\bibliographystyle{ieee_fullname}
\bibliography{egbib}

\begin{thebibliography}{10}\itemsep=-1pt

\bibitem{wgan}
Martin Arjovsky, Soumith Chintala, and L{\'e}on Bottou.
\newblock Wasserstein generative adversarial networks.
\newblock In {\em International Conference on Machine Learning}, pages
  214--223, 2017.

\bibitem{baltruvsaitis2018multimodal}
Tadas Baltru{\v{s}}aitis, Chaitanya Ahuja, and Louis-Philippe Morency.
\newblock Multimodal machine learning: A survey and taxonomy.
\newblock {\em IEEE transactions on pattern analysis and machine intelligence},
  41(2):423--443, 2018.

\bibitem{PPA-cur-2020}
Yidong Chai, Hongyan Liu, and Jie Xu.
\newblock A new convolutional neural network model for peripapillary atrophy
  area segmentation from retinal fundus images.
\newblock {\em Applied Soft Computing}, 86:105890, 2020.

\bibitem{AD-RNN-long-2019-CMIG}
Ruoxuan Cui, Manhua Liu, Alzheimer's Disease~Neuroimaging Initiative, et~al.
\newblock Rnn-based longitudinal analysis for diagnosis of alzheimer’s
  disease.
\newblock {\em Computerized Medical Imaging and Graphics}, 73:1--10, 2019.

\bibitem{AD-MLP-BGRU-2018-ISBI}
Ruoxuan Cui, Manhua Liu, and Gang Li.
\newblock Longitudinal analysis for alzheimer's disease diagnosis using rnn.
\newblock In {\em 2018 IEEE 15th International Symposium on Biomedical Imaging
  (ISBI 2018)}, pages 1398--1401. IEEE, 2018.

\bibitem{who-ref-2019}
Ellen~BM Elsman, Mo Al~Baaj, Gerardus~HMB van Rens, Wencke Sijbrandi, Ellen~GC
  van~den Broek, Hilde~PA van~der Aa, Wouter Schakel, Martijn~W Heymans, Ralph
  de Vries, Mathijs~PJ Vervloed, et~al.
\newblock Interventions to improve functioning, participation, and quality of
  life in children with visual impairment: a systematic review.
\newblock {\em survey of ophthalmology}, 64(4):512--557, 2019.

\bibitem{erhan2010does}
Dumitru Erhan, Aaron Courville, Yoshua Bengio, and Pascal Vincent.
\newblock Why does unsupervised pre-training help deep learning?
\newblock In {\em Proceedings of the thirteenth international conference on
  artificial intelligence and statistics}, pages 201--208, 2010.

\bibitem{goodfellow2014generative}
Ian Goodfellow, Jean Pouget-Abadie, Mehdi Mirza, Bing Xu, David Warde-Farley,
  Sherjil Ozair, Aaron Courville, and Yoshua Bengio.
\newblock Generative adversarial nets.
\newblock In {\em Advances in neural information processing systems}, pages
  2672--2680, 2014.

\bibitem{resnet}
Kaiming He, Xiangyu Zhang, Shaoqing Ren, and Jian Sun.
\newblock Deep residual learning for image recognition.
\newblock In {\em Proceedings of the IEEE conference on computer vision and
  pattern recognition}, pages 770--778, 2016.

\bibitem{hochreiter1997long}
Sepp Hochreiter and J{\"u}rgen Schmidhuber.
\newblock Long short-term memory.
\newblock {\em Neural computation}, 9(8):1735--1780, 1997.

\bibitem{holden2016global}
Brien~A Holden, Timothy~R Fricke, David~A Wilson, Monica Jong, Kovin~S Naidoo,
  Padmaja Sankaridurg, Tien~Y Wong, Thomas~J Naduvilath, and Serge Resnikoff.
\newblock Global prevalence of myopia and high myopia and temporal trends from
  2000 through 2050.
\newblock {\em Ophthalmology}, 123(5):1036--1042, 2016.

\bibitem{kim2018longitudinal}
Martha Kim et~al.
\newblock Longitudinal changes of optic nerve head and peripapillary structure
  during childhood myopia progression on oct: Boramae myopia cohort study
  report 1.
\newblock {\em Ophthalmology}, 125(8):1215--1223, 2018.

\bibitem{kim2012optic}
Tae-Woo Kim, Martha Kim, Robert~N Weinreb, Se~Joon Woo, Kyu~Hyung Park, and
  Jeong-Min Hwang.
\newblock Optic disc change with incipient myopia of childhood.
\newblock {\em Ophthalmology}, 119(1):21--26, 2012.

\bibitem{li2020automatic}
Hanxiang Li et~al.
\newblock Automatic detection of parapapillary atrophy and its association wif
  children myopia.
\newblock {\em Computer methods and programs in biomedicine}, 183:105090, 2020.

\bibitem{follow-up-ipmi-2019}
Maxime Louis, Raphael Couronne, Igor Koval, Benjamin Charlier, and Stanley
  Durrleman.
\newblock Riemannian geometry learning for disease progression modelling.
\newblock In {\em International Conference on Information Processing in Medical
  Imaging}, pages 542--553. Springer, 2019.

\bibitem{lu2011quantification}
Cheng-Kai Lu, Tong~Boon Tang, Augustinus Laude, Ian~J Deary, Baljean Dhillon,
  and Alan~F Murray.
\newblock Quantification of parapapillary atrophy and optic disc.
\newblock {\em Investigative ophthalmology \& visual science},
  52(7):4671--4677, 2011.

\bibitem{PPA-cur-2012}
Cheng-Kai Lu, Tong~Boon Tang, Augustinus Laude, Baljean Dhillon, and Alan~F
  Murray.
\newblock Parapapillary atrophy and optic disc region assessment (pandora):
  retinal imaging tool for assessment of the optic disc and parapapillary
  atrophy.
\newblock {\em Journal of biomedical optics}, 17(10):106010, 2012.

\bibitem{who-2004}
Serge Resnikoff, Donatella Pascolini, Daniel Etya'Ale, Ivo Kocur, Ramachandra
  Pararajasegaram, Gopal~P Pokharel, and Silvio~P Mariotti.
\newblock Global data on visual impairment in the year 2002.
\newblock {\em Bulletin of the world health organization}, 82:844--851, 2004.

\bibitem{OCT-treat-Dense-RNN-long-2020-JBHI}
David~Edmundo Romo-Bucheli, Ursula Schmidt-Erfurth, and Hrvoje Bogunovic.
\newblock End-to-end deep learning model for predicting treatment requirements
  in neovascular amd from longitudinal retinal oct imaging.
\newblock {\em IEEE Journal of Biomedical and Health Informatics}, 2020.

\bibitem{UNet-2D}
Olaf Ronneberger, Philipp Fischer, and Thomas Brox.
\newblock U-net: Convolutional networks for biomedical image segmentation.
\newblock In {\em International Conference on Medical image computing and
  computer-assisted intervention}, pages 234--241. Springer, 2015.

\bibitem{AD-follow-up-baseline-2015}
Daniel Schmitter, Alexis Roche, B{\'e}n{\'e}dicte Mar{\'e}chal, Delphine Ribes,
  Ahmed Abdulkadir, Meritxell Bach-Cuadra, Alessandro Daducci, Cristina
  Granziera, Stefan Kl{\"o}ppel, Philippe Maeder, et~al.
\newblock An evaluation of volume-based morphometry for prediction of mild
  cognitive impairment and alzheimer's disease.
\newblock {\em NeuroImage: Clinical}, 7:7--17, 2015.

\bibitem{song2018progressive}
Min~Kyung Song, Kyung~Rim Sung, Joong~Won Shin, Junki Kwon, Ji~Yun Lee, and
  Ji~Min Park.
\newblock Progressive change in peripapillary atrophy in myopic glaucomatous
  eyes.
\newblock {\em British Journal of Ophthalmology}, 102(11):1527--1532, 2018.

\bibitem{teng2010beta}
Christopher~C Teng, Carlos Gustavo~V De~Moraes, Tiago~S Prata, Celso Tello,
  Robert Ritch, and Jeffrey~M Liebmann.
\newblock $\beta$-zone parapapillary atrophy and the velocity of glaucoma
  progression.
\newblock {\em Ophthalmology}, 117(5):909--915, 2010.

\bibitem{follow-up-AD-2018}
Xiaoqian Wang, Weidong Cai, Dinggang Shen, and Heng Huang.
\newblock Temporal correlation structure learning for mci conversion
  prediction.
\newblock In {\em International Conference on Medical Image Computing and
  Computer-Assisted Intervention}, pages 446--454. Springer, 2018.

\bibitem{Res-2019}
Jianpeng Zhang, Yutong Xie, Yong Xia, and Chunhua Shen.
\newblock Attention residual learning for skin lesion classification.
\newblock {\em IEEE transactions on medical imaging}, 38(9):2092--2103, 2019.

\bibitem{zhang2012automatic}
Zhuo Zhang et~al.
\newblock Automatic glaucoma diagnosis with mrmr-based feature selection.
\newblock {\em J Biomet Biostat S}, 7:2, 2012.

\end{thebibliography}
}
\end{document}